\documentclass{article}

\usepackage{algorithm, algpseudocode}
\usepackage{amsfonts, amsmath, amsopn, amsthm, epsfig}
  \numberwithin{equation}{section}
\usepackage{graphicx, epstopdf}
\usepackage{hyperref}
\usepackage{subfigure}
\usepackage{appendix}

\usepackage{calc}
\usepackage[top=2.5cm,bottom=2.3cm,inner=2.3cm,outer=2.3cm,bindingoffset=0.5cm,footskip=0.55cm]{geometry}
\theoremstyle{remark}

\newenvironment{lemma*}[2][Lemma]{\par\bgroup{\bfseries #1\ #2. }\it\ignorespaces}{\egroup}

\title{Multi-view (Joint) Probability Linear Discrimination Analysis for Multi-view Feature Verification}

\author{Ziqiang Shi\footnotemark[1] \footnotemark[2], Liu Liu\footnotemark[1], Mengjiao Wang\footnotemark[1], Rujie Liu\footnotemark[1]}

\newcommand{\BALD}{\begin{aligned}}
\newcommand{\EALD}{\end{aligned}}
\newcommand{\BALDS}{\begin{aligned*}}
\newcommand{\EALDS}{\end{aligned*}}
\newcommand{\BCAS}{\begin{cases}}
\newcommand{\ECAS}{\end{cases}}
\newcommand{\BEAS}{\begin{eqnarray*}}
\newcommand{\EEAS}{\end{eqnarray*}}
\newcommand{\BEQ}{\begin{equation}}
\newcommand{\EEQ}{\end{equation}}
\newcommand{\BIT}{\begin{itemize}}
\newcommand{\EIT}{\end{itemize}}
\newcommand{\BMAT}{\begin{bmatrix}}
\newcommand{\EMAT}{\end{bmatrix}}
\newcommand{\BNUM}{\begin{enumerate}}
\newcommand{\ENUM}{\end{enumerate}}

\newcommand{\BA}{\begin{array}}
\newcommand{\EA}{\end{array}}

\begin{document}
\date{}
\maketitle

\renewcommand{\thefootnote}{\fnsymbol{footnote}}

\footnotetext[1]{Fujitsu Research \& Development Center, Beijing, China.}
\footnotetext[2]{shiziqiang@cn.fujitsu.com}

\renewcommand{\thefootnote}{\arabic{footnote}}

\begin{abstract}
Multi-view feature has been proved to be very effective in many multimedia applications. However, the current back-end classifiers cannot make full use of such features. In this paper, we propose a method to model the multi-faceted information in the multi-view features explicitly and jointly. In our approach, the feature was modeled as a result derived by a generative multi-view (joint\footnotemark[1]) Probability Linear Discriminant Analysis (PLDA) model, which contains multiple kinds of latent variables. The usual PLDA model only considers one single label. However, in practical use, when using multi-task learned network as feature extractor, the extracted feature are always attached to several labels. This type of feature is called multi-view feature.
With multi-view (joint) PLDA, we are able to explicitly build a model that can combine multiple heterogeneous information from the multi-view features.
In verification step, we calculated the likelihood to describe whether the two features having consistent labels or not. This likelihood are used in the following decision-making. Experiments have been conducted on large scale verification task.
On the public RSR2015 data corpus, the results showed that our approach can achieve 0.02\% EER and 0.09\% EER for impostor wrong and impostor correct cases respectively.

\footnotetext[1]{The term ``joint" comes from the work~\cite{ferrer2017joint} of Dr. Ferrer, who propose a similar idea independently almost at the same time, although they have not do experiments to verify the idea in any applications yet. For this, we will not distinguish between the term ``joint" and ``multi-view" for the description of PLDA in this work}
\end{abstract}

\section{Introduction}

There is a long tradition of using probability linear dimensionality reduction methods for object recognition, for example face recognition and speaker verification~\cite{Ioffe2006Probabilistic,Prince2007Probabilistic,Jiang2012PLDA}. Most notably, these include Factor Analysis (FA) and Probabilistic
Linear Discriminant Analysis (PLDA). While FA captures the main correlations between the coordinates of a feature vector, which may represents
a face image or a speech utterance, PLDA split the total data variability into within-individual and between-individual variabilities,
both residing on small-dimensional subspaces. This makes PLDA suitable for a wide variety of recognition tasks, such as ``one-shot learning''~\cite{Li2003A}, verification (hypothesis testing)~\cite{Prince2007Probabilistic,Variani2014Deep} and etc..

Besides the original PLDA formulation~\cite{Ioffe2006Probabilistic,Prince2007Probabilistic}, there are several alternative
variants:  Gaussian PLDA~\cite{Garcia2011Analysis} and heavy-tailed PLDA~\cite{Kenny2010Bayesian} assume the priors on the model's latent variables follow a Gaussian distribution or Student's t distribution; mixture of PLDA~\cite{simonchik2012supervized,mak2016mixture} assumes the feature is generated from a mixture of factor analysis models.

One of the questions these PLDA methods do not answer is: what shall we do with the situation that the training samples potentially belongs to not only single category, but have different kinds of labels, for example multi-task learned features? Take pose dependent face recognition or text dependent speaker verification for example, the pose or text latent variable is no longer dependent only on the current label, but rather depends on a separate pose or text label. This means the two latent variables related to speaker/individual and pose/text have the equal importance, and both variables are tied across all samples sharing a certain label.

In order to solve this problem, we propose a generalization of the PLDA model called multi-view (joint) PLDA explicitly model jointly multi-view information from samples. The relationship between multi-view (joint) PLDA and standard PLDA is analogous to that between joint factor analysis and factor analysis. Here we need to clarify that the term ``joint PLDA" also used in the work of~\cite{chen2015multi}, however indeed Chen et al. only employed the traditional PLDA but to define the class as the ``joint" classes considering  both speaker and text phrase information.

The paper is organized as follows: sections 2 reviews the traditional PLDA method. Section 3 describe the proposed multi-view (joint) PLDA approach. Section 4 shows real life application of the multi-view (joint) PLDA, the experiments and results of the various systems are discussed. Finally, we conclude in section 5. Full derivation of the EM algorithm and scoring function of the multi-view (joint) PLDA are provided in the appendix of this paper.

\section{Probability Linear Discrimination Analysis}

Probabilistic Linear Discriminant Analysis (PLDA) can be thought of as Linear Discriminant Analysis (LDA) with a probability distributions attached to the features, where the probability distribution models the data through the latent variables corresponding to the class and the view~\cite{Prince2007Probabilistic}.

Take the speaker verification or face recognition for example, assume given the training data consists of $I$ individuals each with $H_i$ utterances or images. PLDA models data generation using the following equation:
\begin{equation}
x_{ij} = \mu + \mathbf{B}z_i + \epsilon_{ij}. \nonumber
\end{equation}
$\epsilon_{ij}$ is defined to be Gaussian with diagonal covariance $\Sigma$. Let $\theta=\{ \mu, \mathbf{B}, \Sigma\}$, $x_i=\{x_{ij}: j=1,...,H_i\}$ and
$X=\{x_{ij}: i=1,...,I;j=1,...,H_i\}$.
More formally the model can be described in terms of conditional probabilities:
\begin{eqnarray}
p(x_{ij}|z_i,\theta) &=& \mathcal{N}(x_{ij}|\mu + \mathbf{B}z_i, \Sigma),  \nonumber\\
p(z_i) &=&  \mathcal{N}(z_i|0,\textbf{I}), \nonumber
\end{eqnarray}
where $\mathcal{N}(x|\mu,\Sigma)$ represents a Gaussian in $x$ with mean $\mu$ and covariance $\Sigma$.

The parameters $\theta$ of this PLDA model can be estimated using the Expectation Maximization (EM)~\cite{Dempster1977Maximum} algorithm.
With the learned PLDA model, given a test $x_t$ and an enrolled model $x_s$, the likelihood ratio score is
\begin{eqnarray*}
l(x_t,x_s) &=& \frac{P(x_t,x_s|\text{same-individual})}{P(x_t,x_s|\text{different-individuals})}\\
&=&\frac{\int p(x_t,x_s,z|\theta)dz}{\int p(x_t,z_t|\theta)dz_t\int p(x_s,z_s|\theta)dz_s} \\
&=& \frac{\int p(x_t,x_s|z,\theta)p(z)dz}{\int p(x_t|z_t,\theta)p(z_t)dz_t\int p(x_s|z_s,\theta)p(z_s)dz_s}\\
 &=&\frac{\mathcal{N}(\left[\begin{matrix}
    x_t\\ x_s
    \end{matrix}\right]|\left[\begin{matrix}
    \mu\\ \mu
    \end{matrix}\right],\left[\begin{matrix}
    \textbf{B}\textbf{B}^T + \Sigma & \textbf{B}\textbf{B}^T\\\textbf{B}\textbf{B}^T & \textbf{B}\textbf{B}^T+\Sigma
    \end{matrix}\right])}{\mathcal{N}(x_t|\mu, \textbf{B}\textbf{B}^T+\Sigma)\mathcal{N}(x_s|\mu, \textbf{B}\textbf{B}^T+\Sigma)}.
\end{eqnarray*}

This standard PLDA cannot properly deal with multi-view features that belong to multiple classes, e.g. the multi-task learning features jointly having different kinds of labels. Typical examples include text dependent speaker verification or pose dependent face recognition.
It is noted that we need to define the PLDA latent variable $z_i$ as the joint variable considering multiple category information. This means the latent variable $z_i$ is dependent on multiple category labels. In this work we try to separate the $z_i$ into independent latent variables - each related to the single category information.
This intuitive idea result in the following multi-view (joint) PLDA.

\section{Multi-view (joint) PLDA}

In this section, we propose an effective method to describe the multi-view features as resulting from a generative model which incorporates jointly both within-multi-view and between-multi-view variation. In verification we calculate the likelihoods that the two vectors having all views consistent or not, and the ratio of these two likelihoods will be used for the final decision.

\subsection{Generative model}

In this work, for notation simplicity we assume the feature has only two kinds of views $\mathcal{A}$ and $\mathcal{B}$, and it can be readily generalized to situations with three or more views.
Let $\theta=\{\mu, \mathbf{S}, \mathbf{T}, \Sigma\}$. We assume that the training (development) data consists of $I$ of $\mathcal{A}$ views (for example speaker/individual identities) and $J$ of $\mathcal{B}$ views (for example phrases/poses) each pair with $H_{ij}$ features. We denote the $k$'th feature with the $i$'th $\mathcal{A}$ view and $j$'th B view by $x_{ijk}$. We model the multi-view feature generation by the process:
\begin{equation}
\label{eq:jplda}
x_{ijk} = \mu + \mathbf{S}u_i + \mathbf{T}v_{j} + \epsilon_{ijk}.
\end{equation}

The model comprises two parts: 1, the signal component $\mu + \mathbf{S}u_i + \mathbf{T}v_{j}$  which depends only on both the $\{\mathcal{A}$,$\mathcal{B}\}$ views but not the particular feature vector (i.e. there is no dependence on $k$); 2, the noise component $\epsilon_{ijk}$ which is different for every feature vector and represents within-multi-view noise. The term $\mu$ represents the overall mean of the training vectors. The columns of the matrix $\mathbf{S}$  and $\mathbf{T}$  contain a basis for the between-$\mathcal{A}$ and between-$\mathcal{B}$ subspaces respectively, while the terms $u_i$ and $v_{j}$ represent the position in these spaces. Remaining unexplained data variation is explained by the residual noise term $\epsilon_{ijk}$ which is defined to be Gaussian with diagonal covariance $\Sigma$.
The latent variables $u_i$ and $v_{j}$ are particularly important in real application, e.g. text dependent speaker verification or pose dependent face recognition, as these represents the identity of the speaker/individual $i$  and the content/angle of the text/pose $j$ respectively. In verification, we will consider the likelihood that the two vectors were generated from the same underlying  $u_i$ and $v_{j}$.

Formally the model can be described in terms of conditional probabilities
\begin{eqnarray*}
p(x_{ijk}|u_i,v_j, \theta) &=& \mathcal{N}(x_{ijk}| \mu + \mathbf{S}u_i + \mathbf{T}v_{j}, \Sigma), \\
p(u_i) &=&  \mathcal{N}(u_i|0,\textbf{I}), \\
p(v_j) &=&  \mathcal{N}(v_j|0,\textbf{I}),
\end{eqnarray*}
where $\mathcal{N}(x|\mu,\Sigma)$ represents a Gaussian in $x$ with mean $\mu$ and covariance $\Sigma$.

Here it's worth to notice that the formulation~\ref{eq:jplda} is almost same as that of Joint Factor Analysis (JFA)~\cite{kenny2005joint}, although the usage is totally different. JFA is mainly used as a front end for robust speaker feature extraction, while in this work the similar formulation is used as back-end for classification of j-vectors. The mathematical relationship between multi-view (joint) PLDA and JFA is analogous (not exactly) to that between PLDA and i-vector~\cite{Jiang2012PLDA}.

Let $X=\{x_{ijk}: i=1,...,I;j=1,...,J;k=1,...,H_{ij}\}$.
In order to find the parameters $\theta=\{\mu, \mathbf{S}, \mathbf{T}, \Sigma\}$ under which the data set $X$ is most likely, the classical EM algorithm~\cite{Dempster1977Maximum} is employed.

\subsection{EM formulation}
 The auxiliary function for EM is
\begin{eqnarray*}
& &Q(\theta|\theta_t) =   \text{E}_{U,V|X,\theta_t}[\log p(X,U,V|\theta) ] \\
&=&\text{E}_{U,V|X,\theta_t}\left\{\sum_{i=1}^I\sum_{j=1}^{J}\sum_{k=1}^{H_{ij}}\log [p(x_{ijk}|u_{i},v_{j}, \theta)p(u_i,v_j)] \right\} \\
&=&\text{E}_{U,V|X,\theta_t}\left\{\sum_{i=1}^I\sum_{j=1}^{J}\sum_{k=1}^{H_{ij}}\log [ \mathcal{N}(x_{ijk}|\mu + \mathbf{S}u_i+ \mathbf{T}v_j, \Sigma)  \mathcal{N}(u_i,v_j|0,\textbf{I})] \right\}
\end{eqnarray*}

Let $z_{ij}=\left[\begin{matrix}
    u_i \\ v_j
    \end{matrix}\right]$, $Z=\{z_{ij}: i=1,...,I;j=1,...,J\}$ and $\mathbf{B}=\left[\begin{matrix}
    \mathbf{S} & \mathbf{T}
    \end{matrix}\right]$.
By maximizing the auxiliary function, we obtain the following EM formulations.

\textbf{E} steps: we need to calculate the expectations $\text{E}_{U|X,\theta_t}[u_i]$, $\text{E}_{V|X,\theta_t}[v_j]$, $\text{E}_{U|X,\theta_t}[u_iu_i^T]$, $\text{E}_{V|X,\theta_t}[v_jv_j^T]$, and  $\text{E}_{U,V|X,\theta_t}[u_iv_j^T]$.
Indeed we have
\begin{eqnarray}
 \label{eq:estep1}
\text{E}_{Z|X,\theta_t}\left\{\left[\begin{matrix}
    u_i \\ v_j
    \end{matrix}\right]\right\} &=&  \\ \left(\textbf{I}+\mathbf{B}^T{\Sigma}^{-1}\mathbf{B}\right)^{-1}&\mathbf{B}^T&{\Sigma}^{-1}\sum_{k=1}^{H_{ij}}(x_{ijk}-\mu), \nonumber
\end{eqnarray}
\begin{eqnarray}
 \label{eq:estep2}
\text{E}_{Z|X,\theta_t}\left\{\left[\begin{matrix}
    u_iu_i^T & u_iv_j^T \\ v_ju_i^T & v_jv_j^T
    \end{matrix}\right]\right\}&=&\left(\textbf{I}+\mathbf{B}^T{\Sigma}^{-1}\mathbf{B}\right)^{-1}\\
    &+&\text{E}_{Z|X,\theta_t}[z_{ij}]\text{E}_{Z|X,\theta_t}[z_{ij}]^T. \nonumber
\end{eqnarray}

\textbf{M} steps: we update the values of the parameters $\theta=\{\mu, \mathbf{S}, \mathbf{T}, \Sigma\}$ and have
\begin{eqnarray*}
\mathbf{S}&=&\{\sum_{i=1}^I\sum_{j=1}^{J}\sum_{k=1}^{H_{ijk}} (x_{ijk}-\mu)\text{E}_{U|X,\theta_t}[u_i]^T
\\&-&\sum_{i=1}^I\sum_{j=1}^{J}\sum_{k=1}^{H_{ijk}} \mathbf{T}\text{E}_{U,V|X,\theta_t}[v_ju_i^T]\}\left\{\text{E}_{U|X,\theta_t}[u_iu_i^T]\right\}^{-1},
\end{eqnarray*}
\begin{eqnarray*}
\mathbf{T}&=&\{\sum_{i=1}^I\sum_{j=1}^{J}\sum_{k=1}^{H_{ijk}} (x_{ijk}-\mu)\text{E}_{V|X,\theta_t}[v_j]^T
\\&-&\sum_{i=1}^I\sum_{j=1}^{J}\sum_{k=1}^{H_{ijk}}\mathbf{S}\text{E}_{U,V|X,\theta_t}[u_iv_j^T]\}\left\{\text{E}_{V|X,\theta_t}[v_jv_j^T]\right\}^{-1},
\end{eqnarray*}
\begin{eqnarray*}
 \Sigma &=&\frac{1}{\sum_{i=1}^I\sum_{j=1}^{J}\sum_{k=1}^{H_{ijk}}1} \sum_{i=1}^I\sum_{j=1}^{J}\sum_{k=1}^{H_{ijk}}  \\
 & &\textbf{diag} \{(x_{ijk}-\mu)(x_{ijk}-\mu)^T \nonumber
 \\& &-(x_{ijk}-\mu)\left[\text{E}_{U|X,\theta_t}[u_i]^T\mathbf{S}^T+\text{E}_{V|X,\theta_t}[v_i]^T\mathbf{T}^T\right]\} \nonumber,
\end{eqnarray*}
\begin{equation*}
\mu = \frac{\sum_{i=1}^I\sum_{j=1}^{J}\sum_{k=1}^{H_{ijk}}  x_{ijk}}{\sum_{i=1}^I\sum_{j=1}^{J}\sum_{k=1}^{H_{ijk}}  1},
\end{equation*}
where \textbf{diag} represents the operation of retaining only the diagonal elements from a matrix.  The expectation terms $\text{E}_{U|X,\theta_t}[u_i]$, $\text{E}_{V|X,\theta_t}[v_j]$, $\text{E}_{U|X,\theta_t}[u_iu_i^T]$, $\text{E}_{V|X,\theta_t}[v_jv_j^T]$, and  $\text{E}_{U,V|X,\theta_t}[u_iv_j^T]$ can be extracted from Equations~\eqref{eq:estep1} and~\eqref{eq:estep2}.

\subsection{Likelihood Ration Scores}

\begin{figure}[!h]
\centering
\includegraphics[width=0.7\textwidth]{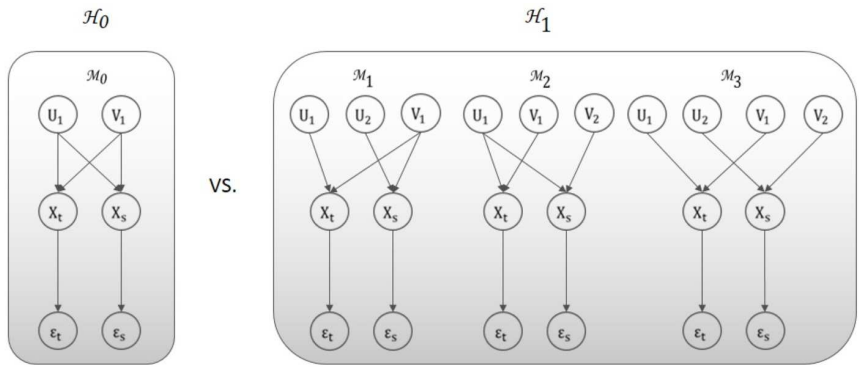}
\caption{Verification by comparing the likelihood of the data under different hypotheses. Under the null hypothesis $\mathcal{H}_1$, the feature $x_t$ and $x_s$ do not match. Under the hypothesis $\mathcal{H}_0$ they match.}
\label{fig:mvPLDA}       
\end{figure}

We treat the verification as a kind of hypothesis testing problem with the null hypothesis $\mathcal{H}_0$ defined for the all same latent variables, and the alternative hypothesis $\mathcal{H}_1$ defined for one or more different latent variables.
That is in verification we compare the likelihood of the vectors under the hypothesis $\mathcal{H}_0$ where two multi-view features match (have the same underlying hidden variables) and the hypothesis $\mathcal{H}_1$ where they do not (different underlying $\mathcal{A}$ variable with same $\mathcal{B}$ variable in model $\mathcal{M}_1$, same $\mathcal{A}$ variable with different $\mathcal{B}$ variables in model $\mathcal{M}_2$, or different underlying $\mathcal{A}$ variables with different $\mathcal{B}$ variables in model $\mathcal{M}_3$, as the Fig.~\ref{fig:mvPLDA} shows). If the two j-vectors belong to the same speaker saying same phrase, then they must have the same speaker and phrase variables $u_i$ and $v_{j}$; otherwise if the two j-vectors belong to different speakers or saying different phrases, they will have different speaker variable or different phrase variables.
Given a test j-vector $x_t$ and an enrolled j-vector $x_s$, set $\mathbf{A} = \mathbf{S}\mathbf{S}^T+\mathbf{T}\mathbf{T}^T$, $\mathbf{B} = 2\mathbf{S}\mathbf{S}^T+\mathbf{T}\mathbf{T}^T$, $\mathbf{C} = \mathbf{S}\mathbf{S}^T+2\mathbf{T}\mathbf{T}^T$, and let the priori probability of the models $\mathcal{M}_1$, $\mathcal{M}_2$, $\mathcal{M}_3$ as $p_1=P(\mathcal{M}_1|\mathcal{H}_1)$, $p_2=P(\mathcal{M}_2|\mathcal{H}_1)$, $p_3=P(\mathcal{M}_3|\mathcal{H}_1)$, then the likelihood ratio score is
\begin{eqnarray*}
& &l(x_t,x_s)= \frac{P(x_t,x_s|\mathcal{H}_0)}{P(x_t,x_s|\mathcal{H}_1)}\\&=&\frac{\int\int p(x_t,x_s|u_1,v_1,\theta)p(u_1)p(v_1)du_1dv_1}{\textbf{X}} \\
&=&\frac{\mathcal{N}(\left[\begin{matrix}
    x_t\\ x_s
    \end{matrix}\right]|\left[\begin{matrix}
    \mu\\ \mu
    \end{matrix}\right],\left[\begin{matrix}
    \mathbf{A}  + \Sigma & \mathbf{A}   \\ \mathbf{A}   & \mathbf{A}  +\Sigma
    \end{matrix}\right])}{\textbf{X}},
\end{eqnarray*}
where
\begin{eqnarray*}
& &\textbf{X}=P(x_t,x_s|\mathcal{H}_1)\\&=& P(x_t,x_s|\mathcal{M}_1)P(\mathcal{M}_1|\mathcal{H}_1)
\\&+&P(x_t,x_s|\mathcal{M}_2)P(\mathcal{M}_2|\mathcal{H}_1)+P(x_t,x_s|\mathcal{M}_3)P(\mathcal{M}_3|\mathcal{H}_1)\\&+&p_1\int\int\int  p(x_t,x_s,u_1,u_2,v_1|\theta)du_1du_2dv_1\\&=&p_2\int\int\int  p(x_t,x_s,u_1,v_1,v_2|\theta)du_1dv_1dv_2\\ &+&p_3\int\int  p(x_t,u_1,v_1|\theta)du_1dv_1\int\int  p(x_s,u_2,v_2|\theta)du_2dv_2 \\
&=&p_1\mathcal{N}(\left[\begin{matrix}
    x_t\\ x_s
    \end{matrix}\right]|\left[\begin{matrix}
    \mu\\ \mu
    \end{matrix}\right],\left[\begin{matrix}
  \mathbf{C}+ \Sigma & \mathbf{C}\\ \mathbf{C}  & \mathbf{C}+\Sigma
    \end{matrix}\right])\\&+&p_2\mathcal{N}(\left[\begin{matrix}
    x_t\\ x_s
    \end{matrix}\right]|\left[\begin{matrix}
    \mu\\ \mu
    \end{matrix}\right],\left[\begin{matrix}
    \mathbf{B}  + \Sigma & \mathbf{B}   \\\mathbf{B}   &\mathbf{B}  +\Sigma
    \end{matrix}\right])\\ &+& p_3\mathcal{N}(x_t|\mu, \mathbf{A} +\Sigma)\mathcal{N}(x_s|\mu, \mathbf{A} +\Sigma).
\end{eqnarray*}

Notice that like standard PLDA~\cite{Prince2007Probabilistic}, we do not calculate a point estimate of hidden variable. Instead we compute the probability that the two multi-view vectors had the same hidden variables, regardless of what this actual latent variable was.

\section{Experiments}

Multi-view (joint) PLDA is a general method for object recognition.
In this section, the method is evaluated on the task of text dependent speaker verification. We describe the experimental setup and results for the proposed method on the public RSR2015 English corpus~\cite{larcher2014text} and our internal Huiting202 Chinese Mandarin database collected by the Huiting Techonogly\footnotemark[2].

\footnotetext[2]{http://huitingtech.com/}

\subsection{J-vector extraction}

\begin{figure}[!h]
\centering
\includegraphics[width=0.6\textwidth]{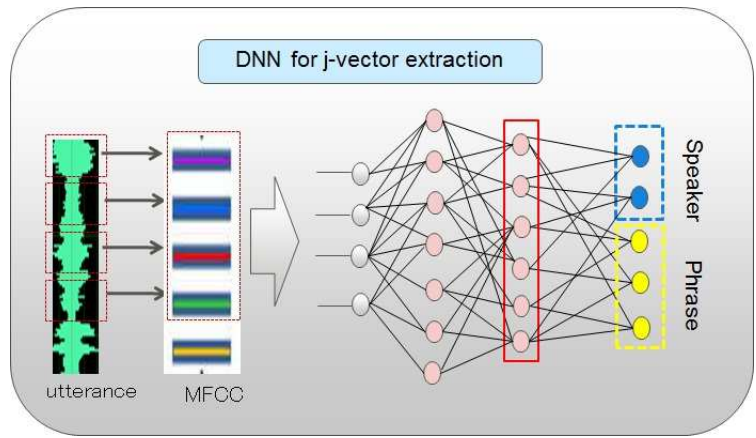}
\caption{Multi-task joint learning DNN as j-vector extractor.}
\label{fig:jvector}       
\end{figure}

Chen et al.~\cite{chen2015multi} proposed a method to train a DNN to make classifications for both speaker and phrase identities by minimizing a total loss function consisting a sum of two cross-entropy losses as shown in Fig.~\ref{fig:jvector} - one related to the speaker label and the other to the text label. Once training is complete, the output layer is removed, and the rest of the neural network is used to extract speaker-phrase joint features. That is each frame of an utterance is forward propagated through the network, and the output activations of all the frames are averaged to form an utterance-level feature called j-vector. The enrollment speaker models are formed by averaging the j-vectors corresponding to the enrollment recordings.

\subsection{Experimental setup}

RSR2015 corpus~\cite{larcher2014text} was released by I2R, is used to evaluate the performance of different speaker verification systems. In this work, we follow the setup of~\cite{liu2015deep}, the part I of RSR2015 is used for the testing of multi-view (joint) PLDA. The background and development data of RSR2015 part I are merged as new background data to train the j-vector extractor.

Our internal Huiting202 database is designed for local applications. It contains 202 speakers reading 20 different phrases, 20 sessions each phrase. All speech files are of 16kHz. The gender distribution is balanced on the data set. 132 randomly selected speakers are used for training the background multi-task learned DNN, and the remaining 70 speakers were used for enrollment and evaluation.

In this work, 39-dimensional Mel-frequency cepstral coefficients (MFCC, 13 static including the
log energy + 13 $\Delta$ + 13 $\Delta \Delta$) are extracted and normalized using utterance-level
mean and variance normalization. The input is stacked normalized MFCCs from 11 frames (5 frames from
each side of the current frame).
The DNN has 6 hidden layers (with sigmoid activation
function) of 2048 nodes each. During the background model development stage, the DNN was trained by the strategy of pre-training with Restricted Boltzmann Machine (RBM) and fine tuning with SGD using cross-entropy criterion.
Once the DNN is trained, the j-vector can be extracted during the enrollment and evaluation stages.

\subsection{Results and discussion}

Three systems are evaluated and compared across above conditions:
\begin{itemize}
\item
\textbf{j-vector}: the standard j-vector system with cosine similarity.
\item
\textbf{PLDA}: the j-vector system with classic PLDA (the j-vector with ``joint PLDA" called in~\cite{chen2015multi}).
\item
\textbf{jPLDA}: multi-view (joint) PLDA system described in section~\ref{sec:mvplda} with j-vector.
\end{itemize}

When evaluation a speaker is enrolled with 3 utterances of the same phrase. The task concerns on both the phrase content and speaker identity.
Nontarget trials are of three types: the impostor pronouncing wrong lexical content (impostor wrong, IW); a target speaker pronouncing wrong lexical content (target wrong, TW); the imposter pronouncing correct lexical content (impostor correct, IC).

The PLDA and jPLDA models both are trained using the j-vectors. The class defined in both models is the multi-task label on both the speaker and phrase. For each test session the j-vector is extracted using the same process and then log likelihood from PLDA and likelihood from jPLDA are used to distinguish among different models. The subspace dimension is set to 40 and then the PLDA model is estimated  with 10 iterations; the speaker and the phrase subspace dimensions of jPLDA are both set to 20 in order for fair comparisons and the jPLDA mode is also trained with 10 iterations.

Table~\ref{tab:rsr2015} and~\ref{tab:huiting202} compare the performances of all above-mentioned systems in terms of equal error rate (EER) for the three types of nontarget trials. Obviously jPLDA is superior thanthe standard PLDA, no matter which database is used in the test.
Since multi-view PLDA system can explore both the identity and the lexical information from the j-vector, it always performs better than conventional PLDA systems.

\begin{table}[th]
\caption[rsr2015]{Performance of different systems on the evaluation set of RSR2015 part I in terms of equal error rate (EER \%).}\label{tab:rsr2015}
\centering
\begin{tabular}{|c|c|c|c|}
\hline
EER(\%) & j-vector &  PLDA & jPLDA \\
\hline
IW & 0.95& 0.03 &   0.02 \\
\hline
IC & 7.86 & 6.50 & 3.23 \\
\hline
TW & 3.14 & 0.11 &  0.09 \\
\hline
Total & 1.45 &0.73 & 0.41  \\
\hline
\end{tabular}
\end{table}

\begin{table}[th]
\caption[huiting202]{Performance of different systems on the evaluation set of Huiting202 in terms of equal error rate (EER \%).}\label{tab:huiting202}
\centering
\begin{tabular}{|c|c|c|c|}
\hline
EER(\%) & j-vector &  PLDA & jPLDA \\
\hline
IW & 0.86& 0.17 &  0.13  \\
\hline
IC & 4.57 & 3.91 & 2.37 \\
\hline
TW & 6.71 & 0.83 &  0.70 \\
\hline
Total & 1.37 & 0.62 &  0.38 \\
\hline
\end{tabular}
\end{table}

\section{Conclusions}

We presented a novel generative model that decomposes a pattern into the different classes and the multiple views. Multi-view (joint) Probabilistic Linear Discriminant Analysis (PLDA) is related to PLDA, and can be thought of as PLDA with multiple probability distributions attached to or influence the features. One of the most important advantages of multi-view PLDA, compared to PLDA and its previously proposed probabilistic motivations, is that multiple information can be explicitly modeled and explored from the samples to improve the verification performance.
 Reported results showed that multi-view PLDA provided significant reduction in error rates over conventional systems in term of EER.

\bibliographystyle{plain}
\bibliography{multiview_plda}

\begin{thebibliography}{10}

\bibitem{chen2015multi}
Nanxin Chen, Yanmin Qian, and Kai Yu.
\newblock Multi-task learning for text-dependent speaker verificaion.
\newblock In {\em INTERSPEECH}, 2015.

\bibitem{Dempster1977Maximum}
A.~P. Dempster.
\newblock Maximum likelihood estimation from incomplete data via the em
  algorithm (with discussion.
\newblock {\em Journal of the Royal Statistical Society}, 39(1):1--38, 1977.

\bibitem{ferrer2017joint}
Luciana Ferrer.
\newblock Joint probabilistic linear discriminant analysis.
\newblock {\em arXiv preprint arXiv:1704.02346}, 2017.

\bibitem{Garcia2011Analysis}
Daniel Garcia-Romero and Carol~Y. Espy-Wilson.
\newblock Analysis of i-vector length normalization in speaker recognition
  systems.
\newblock In {\em INTERSPEECH 2011, Conference of the International Speech
  Communication Association, Florence, Italy, August}, pages 249--252, 2011.

\bibitem{Higham1996Accuracy}
Higham and J~Nicholas.
\newblock Accuracy and stability of numerical algorithms.
\newblock {\em Journal of the American Statistical Association},
  16(94):285--289, 1996.

\bibitem{Ioffe2006Probabilistic}
Sergey Ioffe.
\newblock Probabilistic linear discriminant analysis.
\newblock {\em Proc Eccv}, 22(4):531--542, 2006.

\bibitem{Jiang2012PLDA}
Y.~Jiang, K.~A. Lee, Z.~Tang, B.~Ma, A.~Larcher, and H.~Li.
\newblock Plda modeling in i-vector and supervector space for speaker
  verification.
\newblock In {\em ACM International Conference on Multimedia, Singapore,
  November}, pages 882--891, 2012.

\bibitem{Kenny2010Bayesian}
P.~Kenny.
\newblock Bayesian speaker verification with heavy tailed priors.
\newblock In {\em Proc. Odyssey Speaker and Language Recogntion Workshop, Brno,
  Czech Republic}, 2010.

\bibitem{larcher2014text}
Anthony Larcher, Kong~Aik Lee, Bin Ma, and Haizhou Li.
\newblock Text-dependent speaker verification: Classifiers, databases and
  rsr2015.
\newblock {\em Speech Communication}, 60:56--77, 2014.

\bibitem{Li2003A}
Fe~Fei Li, R~Fergus, and P~Perona.
\newblock A bayesian approach to unsupervised one-shot learning of object
  categories.
\newblock In {\em IEEE International Conference on Computer Vision, 2003.
  Proceedings}, pages 1134--1141 vol.2, 2003.

\bibitem{liu2015deep}
Yuan Liu, Yanmin Qian, Nanxin Chen, Tianfan Fu, Ya~Zhang, and Kai Yu.
\newblock Deep feature for text-dependent speaker verification.
\newblock {\em Speech Communication}, 73:1--13, 2015.

\bibitem{mak2016mixture}
Man-Wai Mak, Xiaomin Pang, and Jen-Tzung Chien.
\newblock Mixture of plda for noise robust i-vector speaker verification.
\newblock {\em IEEE/ACM Transactions on Audio, Speech and Language Processing
  (TASLP)}, 24(1):130--142, 2016.

\bibitem{Prince2007Probabilistic}
Simon J~D Prince and James~H Elder.
\newblock Probabilistic linear discriminant analysis for inferences about
  identity.
\newblock {\em Proceedings}, pages 1--8, 2007.

\bibitem{simonchik2012supervized}
Konstantin Simonchik, Timur Pekhovsky, Andrey Shulipa, and Anton Afanasyev.
\newblock Supervized mixture of plda models for cross-channel speaker
  verification.
\newblock In {\em INTERSPEECH}, pages 1684--1687, 2012.

\bibitem{Variani2014Deep}
Ehsan Variani, Xin Lei, Erik Mcdermott, and Ignacio~Lopez Moreno.
\newblock Deep neural networks for small footprint text-dependent speaker
  verification.
\newblock In {\em ICASSP 2014 - 2014 IEEE International Conference on
  Acoustics, Speech and Signal Processing}, pages 4052--4056, 2014.

\end{thebibliography}

\appendix

This appendix provides the derivation of the formulae about multi-view (joint) Probability Linear Discriminant Analysis (PLDA). The appendix starts with the general Expectation Maximization (EM) algorithm and EM to the conventional PLDA. Then, the appendix extends the derivation to multi-view PLDA.

\section{The Expectation Maximization (EM) algorithm}

Given the statistical model which generates a set $X$ of observed data, a set of unobserved latent data or missing values $Z$, and a vector of unknown parameters $\theta$, along with a likelihood function $\mathcal{L}(\theta;X,Z)=p(X,Z|\theta)$, the maximum likelihood estimate (MLE) of the unknown parameters is determined by the marginal likelihood of the observed data
\begin{equation*}
\mathcal{L}(\theta;X)=p(X|\theta)=\sum_{Z}p(X,Z|\theta).\nonumber
\end{equation*}

However, this quantity is often intractable.
The EM algorithm seeks to find the MLE of the marginal likelihood by iteratively applying the following two steps:

\textbf{Expectation step (E step)}: Calculate the expected value of the log likelihood function, with respect to the conditional distribution of $Z$  given $X$  under the current estimate of the parameters $\theta_t$
\begin{equation*}
Q(\theta|\theta_t)=\text{E}_{Z|X,\theta_t}[\log \mathcal{L}(\theta;X,Z)]. \nonumber
\end{equation*}

\textbf{Maximization step (M step)}: Find the parameter that maximizes this quantity:
\begin{equation*}
\theta_{t+1} := \arg\max_{\theta}Q(\theta|\theta_t).\nonumber
\end{equation*}

Expectation-maximization works to improve $Q({ {\theta }}|{ {\theta }}_t)$ rather than directly improving $ \log p(X|\theta)$. Here we show that improvements to the former imply improvements to the latter.

For any $Z$  with non-zero probability $p(Z|X,\theta)$, we can write
$\log p(X|\theta) = \log p(X,Z|\theta) - \log p(Z|X,\theta)$\,.
We take the expectation over possible values of the unknown data $Z$  under the current parameter estimate $ \theta_t$ by multiplying both sides by $p(Z|X,\theta_t)$ and summing (or integrating) over $Z$. The left-hand side is the expectation of a constant, so we get:

\begin{align*}
\log p(X|\theta) &
= \sum_{Z} p(Z|X,\theta_t) \log p(X,Z|\theta)
- \sum_{Z} p(Z|X,\theta_t) \log p(Z|X,\theta)\nonumber \\
& = Q(\theta|\theta_t) + H(\theta|\theta_t) \,,\nonumber
\end{align*}
where $H(\theta|\theta_t)$ is defined by the negated sum it is replacing. This last equation holds for any value of $\theta$ including $\theta = \theta_t$,
$\log p(X|\theta_t)
= Q(\theta_t|\theta_t) + H(\theta_t|\theta_t)\,,$
and subtracting this last equation from the previous equation gives
$\log p(X|\theta) - \log p(X|\theta_t)
= Q(\theta|\theta_t) - Q(\theta_t|\theta_t)
 + H(\theta|\theta_t) - H(\theta_t|\theta_t) \,,$
However, Gibbs' inequality tells us that $ H(\theta|\theta_t) \ge H(\theta_t|\theta_t)$, so we can conclude that

$\log p(X|\theta) - \log p(X|\theta_t)
\ge Q(\theta|\theta_t) - Q(\theta_t|\theta_t) \,.$
In words, choosing ${ {\theta }}$ to improve $Q(\theta|\theta_t)$ beyond $Q(\theta_t|\theta_t)$ can not cause $\log p(X|\theta)$ to decrease below $\log p(X|\theta_t)$, and so the marginal likelihood of the data is non-decreasing.

Observations $X = \{x_1,...,x_n\}$, latent variables $Z = \{z_1,...,z_m\}$, the model $p(x|\theta)$.

\begin{equation*}\label{eq:likelihood}
\mathcal{L}(\theta)=\sum_{i=1}^n \log p(x_i|\theta)=\sum_{i=1}^n \log \sum_{j=1}^m p(x_i,z_j|\theta).
\end{equation*}

\begin{eqnarray*}
\sum_{i=1}^n \log p(x_i|\theta) &=& \sum_{i=1}^n \log \sum_{j=1}^m p(x_i,z_j|\theta) \\
 &=&  \sum_{i=1}^n \log \sum_{j=1}^m Q_i(z_j)\frac{p(x_i,z_j|\theta)}{Q_i(z_j)} \\
 &\geq& \sum_{i=1}^n \sum_{j=1}^m Q_i(z_j) \log \frac{p(x_i,z_j|\theta)}{Q_i(z_j)}.
\end{eqnarray*}
where $\sum_{j=1}^m Q_i(z_j) = 1$, and $\geq$ becomes $=$ if and only if $Q_i(z_j)=p(z_j|x_i, \theta)$.

Thus we get the EM algorithm.

\textbf{E} steps:
\begin{equation*}\label{eq:e_step}
Q_i(z_j):=p(z_j|x_i, \theta_t).
\end{equation*}

\textbf{M} steps:
\begin{equation*}\label{eq:m_step}
\theta_{t+1} := \arg\max_{\theta}\sum_{i=1}^n \sum_{j=1}^m Q_i(z_j) \log \frac{p(x_i,z_j|\theta)}{Q_i(z_j)}.
\end{equation*}

The convergence of EM algorithm.
We have
\begin{equation*}\label{eq:likelihood_t}
\mathcal{L}(\theta_t)= \sum_{i=1}^n \sum_{j=1}^m  Q_i(z_j) \log \frac{p(x_i,z_j|\theta_t)}{Q_i(z_j)},
\end{equation*}
where $Q_i(z_j):=p(z_j|x_i, \theta_t)$.

\begin{eqnarray*}
\mathcal{L}(\theta_{t+1}) &=& \max_{\theta}\sum_{i=1}^n \sum_{j=1}^m Q_i(z_j) \log \frac{p(x_i,z_j|\theta)}{Q_i(z_j)} \\
 &\geq&  \sum_{i=1}^n \sum_{j=1}^m Q_i(z_j) \log \frac{p(x_i,z_j|\theta_t)}{Q_i(z_j)}   \\
 &=& \mathcal{L}(\theta_t).
\end{eqnarray*}

\section{Probability Linear Discriminant Analysis}

We assume that the training data consists of $I$ speakers or individuals each with $H_i$ sessions or images. We model data generation by the process:
\begin{equation}
x_{ij} = \mu + \mathbf{B}z_i + \epsilon_{ij}. \nonumber
\end{equation}
$\epsilon_{ij}$ is defined to be Gaussian with diagonal covariance $\Sigma$. Let $\theta=\{ \mu, \mathbf{B}, \Sigma\}$, $x_i=\{x_{ij}: j=1,...,H_i\}$ and
$X=\{x_{ij}\in \mathbb{R}^d: i=1,...,I;j=1,...,H_i\}$ and $\mathbf{B}\in \mathbb{R}^{d\times N}$, that is the dimension of the subspace is $N$.

More formally the model can be described in terms of conditional probabilities:
\begin{eqnarray}
P(x_{ij}|z_i,\theta) &=& \mathcal{N}(x_{ij}|\mu + \mathbf{B}z_i, \Sigma),  \nonumber\\
P(z_i) &=&  \mathcal{N}(z_i|0,\textbf{I}), \nonumber
\end{eqnarray}
where $\mathcal{N}(x|\mu,\Sigma)$ represents a Gaussian in $x$ with mean $\mu$ and covariance $\Sigma$.

\subsection{Training of PLDA}

We use EM algorithm updates for learning the PLDA model.

\textbf{E} steps: calculate the expectation $\text{E}_{Z|X,\theta_t}[z_i]$ and $\text{E}_{Z|X,\theta_t}[z_iz_i^T]$.
\begin{eqnarray*}
P(z_i|x_i,\theta)&\propto& P(x_i|z_i,\theta)P(z_i) = \left[\prod_{j=1}^{H_i}P(x_{ij}|z_i,\theta)\right]P(z_i)\\
&=&\left[\prod_{j=1}^{H_i}\mathcal{N}(x_{ij}|\mu+\mathbf{B}z_i, \Sigma)\right]\mathcal{N}(z_i|0,\textbf{I})\\
&\propto& \left\{\prod_{j=1}^{H_i} \exp \left[-\frac{1}{2}(x_{ij}-\mu-\mathbf{B}z_i)^T {\Sigma}^{-1}(x_{ij}-\mu-\mathbf{B}z_i)\right]\right\}\exp\left(-\frac{1}{2}z_i^Tz_i\right) \\
&=& \exp \left\{ -\frac{1}{2}\left[\sum_{j=1}^{H_i} (x_{ij}-\mu-\mathbf{B}z_i)^T {\Sigma}^{-1}(x_{ij}-\mu-\mathbf{B}z_i)\right] -\frac{1}{2}z_i^Tz_i\right\} \\
&\propto& \exp \left\{ \sum_{j=1}^{H_i} (x_{ij}-\mu)^T {\Sigma}^{-1}\mathbf{B}z_i-\frac{1}{2}z_i^T(\textbf{I}+H_i\mathbf{B}^T{\Sigma}^{-1}\mathbf{B})z_i\right\}
\end{eqnarray*}

Thus we have
\begin{equation*}
\text{E}_{Z|X,\theta_t}[z_i]=\left(\textbf{I}+H_i\mathbf{B}^T{\Sigma}^{-1}\mathbf{B}\right)^{-1}\mathbf{B}^T{\Sigma}^{-1}\sum_{j=1}^{H_i}(x_{ij}-\mu).
\end{equation*}

\begin{equation*}
\text{E}_{Z|X,\theta_t}[z_iz_i^T]=\left(\textbf{I}+H_i\mathbf{B}^T{\Sigma}^{-1}\mathbf{B}\right)^{-1}+\text{E}_{Z|X,\theta_t}[z_i]\text{E}_{Z|X,\theta_t}[z_i]^T.
\end{equation*}

\textbf{M} steps:
\begin{eqnarray*}
\theta_{t+1} &:=& \arg\max_{\theta}Q(\theta|\theta_t) \\
&=&  \arg\max_{\theta} \text{E}_{Z|X,\theta_t}[\log \mathcal{L}(\theta;X,Z)] \\
&=&  \arg\max_{\theta} \text{E}_{Z|X,\theta_t}[\log p(X,Z|\theta) ] \\
&=&  \arg\max_{\theta} \text{E}_{Z|X,\theta_t}\left[\sum_{i=1}^I\sum_{j=1}^{H_i}\log p(x_{ij},z_{i}|\theta) \right] \\
&=&  \arg\max_{\theta} \text{E}_{Z|X,\theta_t}\left\{\sum_{i=1}^I\sum_{j=1}^{H_i}\log [p(x_{ij}|z_{i}, \theta)p(z_i)] \right\} \\
&=& \arg\max_{\theta} \text{E}_{Z|X,\theta_t}\left\{\sum_{i=1}^I\sum_{j=1}^{H_i}\log [ \mathcal{N}(x_{ij}|\mu + \mathbf{B}z_i, \Sigma) \mathcal{N}(z_i|0,\textbf{I})] \right\} \\
&=& \arg\max_{\theta} \text{E}_{Z|X,\theta_t}\left\{\sum_{i=1}^I\sum_{j=1}^{H_i}\log \left[ \frac{1}{\sqrt{(2\pi)^D|\Sigma|}}\exp \left(-\frac{1}{2}(x_{ij}-\mu-\mathbf{B}z_i)^T\Sigma^{-1}(x_{ij}-\mu-\mathbf{B}z_i)\right)\right] \right\} \\
&=& - \arg\max_{\theta} \sum_{i=1}^I\sum_{j=1}^{H_i}\text{E}_{Z|X,\theta_t}\left\{ \frac{1}{2}\log |\Sigma|+ \frac{1}{2}(x_{ij}-\mu-\mathbf{B}z_i)^T\Sigma^{-1}(x_{ij}-\mu-\mathbf{B}z_i)) \right\} \\
&=&\arg\max_{\theta} \sum_{i=1}^I\sum_{j=1}^{H_i}\left[ - \frac{1}{2}\log |\Sigma| - \frac{1}{2}(x_{ij}-\mu)^T\Sigma^{-1}(x_{ij}-\mu)\right] \\& &+ \sum_{i=1}^I\sum_{j=1}^{H_i} (x_{ij}-\mu)^T\Sigma^{-1}\mathbf{B}\text{E}_{Z|X,\theta_t}[z_i] - \frac{1}{2}\left[\sum_{i=1}^I\sum_{j=1}^{H_i}\text{E}_{Z|X,\theta_t}[z_i^T\mathbf{B}^T\Sigma^{-1}\mathbf{B}z_i]\right]
\end{eqnarray*}
where the fact that $\mathcal{N}(z_i|0,\textbf{I})$ has no dependence on $\theta$ is used in the derivation.

Take derivatives with respect to $\mathbf{B}$, $\Sigma^{-1}$, and $\mu$\, and then equate these derivatives to zero to proved the update rules.
The following is the detailed derivations.

We have
\begin{equation*}
\frac{\partial Q}{\partial \mathbf{B}}=\sum_{i=1}^I\sum_{j=1}^{H_i} \Sigma^{-1}(x_{ij}-\mu)\text{E}_{Z|X,\theta_t}[z_i]^T - \sum_{i=1}^I\sum_{j=1}^{H_i} \Sigma^{-1}\mathbf{B}\text{E}_{Z|X,\theta_t}[z_iz_i^T].
\end{equation*}

Setting $\frac{\partial Q}{\partial \mathbf{B}}=0$, result in
\begin{equation*}
\mathbf{B}=\left\{\sum_{i=1}^I\sum_{j=1}^{H_i} (x_{ij}-\mu)\text{E}_{Z|X,\theta_t}[z_i]^T\right\}\left\{\sum_{i=1}^I\sum_{j=1}^{H_i}\text{E}_{Z|X,\theta_t}[z_iz_i^T]\right\}^{-1}.
\end{equation*}

We have
\begin{eqnarray*}
\frac{\partial Q}{\partial \Sigma^{-1}} &=&  \frac{1}{2}\sum_{i=1}^I\sum_{j=1}^{H_i} \left[\Sigma - (x_{ij}-\mu)(x_{ij}-\mu)^T\right] \\
& &+ \sum_{i=1}^I\sum_{j=1}^{H_i} (x_{ij}-\mu)\text{E}_{Z|X,\theta_t}[z_i]^T\mathbf{B}^T - \frac{1}{2}\left[\sum_{i=1}^I\sum_{j=1}^{H_i}\mathbf{B}\text{E}_{Z|X,\theta_t}[z_iz_i^T]\mathbf{B}^T\right] \\
&=&  \frac{1}{2}\sum_{i=1}^I\sum_{j=1}^{H_i} \left[\Sigma - (x_{ij}-\mu)(x_{ij}-\mu)^T\right] \\
& &+ \sum_{i=1}^I\sum_{j=1}^{H_i} (x_{ij}-\mu)\text{E}_{Z|X,\theta_t}[z_i]^T\mathbf{B}^T - \frac{1}{2}\left[\sum_{i=1}^I\sum_{j=1}^{H_i}(x_{ij}-\mu)\text{E}_{Z|X,\theta_t}[z_i]^T\mathbf{B}^T\right]
\end{eqnarray*}

Setting $\frac{\partial Q}{\partial\Sigma^{-1}}=0$, we have
\begin{equation*}
\sum_{i=1}^I\sum_{j=1}^{H_i} \Sigma = \sum_{i=1}^I\sum_{j=1}^{H_i} \left\{(x_{ij}-\mu)(x_{ij}-\mu)^T-(x_{ij}-\mu)\text{E}_{Z|X,\theta_t}[z_i]^T\mathbf{B}^T\right\}.
\end{equation*}

Rearranging, result in
\begin{equation*}
 \Sigma =\frac{1}{\sum_{i=1}^I\sum_{j=1}^{H_i}1} \sum_{i=1}^I\sum_{j=1}^{H_i} \left\{(x_{ij}-\mu)(x_{ij}-\mu)^T-(x_{ij}-\mu)\text{E}_{Z|X,\theta_t}[z_i]^T\mathbf{B}^T\right\}.
\end{equation*}

We have
\begin{equation*}
\frac{\partial Q}{\partial \mu} = \sum_{i=1}^I\sum_{j=1}^{H_i}  - (x_{ij}-\mu)^T\Sigma^{-1} - \sum_{i=1}^I\sum_{j=1}^{H_i} \text{E}_{Z|X,\theta_t}[z_i]^T\mathbf{B}^T\Sigma^{-1}
\end{equation*}
Setting $\frac{\partial Q}{\partial \mu}=0$, we have
\begin{equation*}
\sum_{i=1}^I\sum_{j=1}^{H_i} - (x_{ij}-\mu)^T- \sum_{i=1}^I\sum_{j=1}^{H_i} \text{E}_{Z|X,\theta_t}[z_i]^T\mathbf{B}^T=0
\end{equation*}

Since $\text{E}_{Z|X,\theta_t}[z_i]\approx 0$, we have $\sum_{i=1}^I\sum_{j=1}^{H_i} - (x_{ij}-\mu)=0$, that is
\begin{equation*}
\mu = \frac{\sum_{i=1}^I\sum_{j=1}^{H_i} x_{ij}}{\sum_{i=1}^I\sum_{j=1}^{H_i} 1}.
\end{equation*}

\subsection{Verification by using PLDA}
Given a test $x_t$ and target $x_s$, the likelihood ratio score is
\begin{eqnarray*}
l(x_t,x_s) &=& \frac{P(x_t,x_s|\text{same latent variable})}{P(x_t,x_s|\text{different latent variables})}=\frac{\int p(x_t,x_s,z|\theta)dz}{\int p(x_t,z_t|\theta)dz_t\int p(x_s,z_s|\theta)dz_s} \\
&=& \frac{\int p(x_t,x_s|z,\theta)p(z)dz}{\int p(x_t|z_t,\theta)p(z_t)dz_t\int p(x_s|z_s,\theta)p(z_s)dz_s} =\frac{\mathcal{N}(\left[\begin{matrix}
    x_t\\ x_s
    \end{matrix}\right]|\left[\begin{matrix}
    \mu\\ \mu
    \end{matrix}\right],\left[\begin{matrix}
    \textbf{B}\textbf{B}^T + \Sigma & \textbf{B}\textbf{B}^T\\\textbf{B}\textbf{B}^T & \textbf{B}\textbf{B}^T+\Sigma
    \end{matrix}\right])}{\mathcal{N}(x_t|\mu, \textbf{B}\textbf{B}^T+\Sigma)\mathcal{N}(x_s|\mu, \textbf{B}\textbf{B}^T+\Sigma)}.
\end{eqnarray*}

We take $\log$ of both side
\begin{equation*}
\log l(x_t,x_s) = \log \mathcal{N}(\left[\begin{matrix}
    x_t\\ x_s
    \end{matrix}\right]|\left[\begin{matrix}
    \mu\\ \mu
    \end{matrix}\right],\left[\begin{matrix}
    \textbf{B}\textbf{B}^T + \Sigma & \textbf{B}\textbf{B}^T\\\textbf{B}\textbf{B}^T & \textbf{B}\textbf{B}^T+\Sigma
    \end{matrix}\right]) - \log  \mathcal{N}(x_t|\mu, \textbf{B}\textbf{B}^T+\Sigma)\mathcal{N}(x_s|\mu, \textbf{B}\textbf{B}^T+\Sigma).
\end{equation*}

Since $\mu$ is a global offset that can be pre-computed and removed from all $x_{ij}$, we set $\mu=0$, and we have
\begin{eqnarray*}
\log l(x_t,x_s) &=& -\frac{1}{2}\left[\begin{matrix}
    x_t^T & x_s^T
    \end{matrix}\right]\left[\begin{matrix}
    \textbf{B}\textbf{B}^T + \Sigma & \textbf{B}\textbf{B}^T\\\textbf{B}\textbf{B}^T & \textbf{B}\textbf{B}^T+\Sigma
    \end{matrix}\right]^{-1}\left[\begin{matrix}
    x_t\\ x_s
    \end{matrix}\right]\\&+&\frac{1}{2}\left[\begin{matrix}
     x_t^T & x_s^T
    \end{matrix}\right]\left[\begin{matrix}
    \textbf{B}\textbf{B}^T + \Sigma &0\\0 & \textbf{B}\textbf{B}^T+\Sigma
    \end{matrix}\right]^{-1}\left[\begin{matrix}
    x_t\\ x_s
    \end{matrix}\right]+const.
\end{eqnarray*}

Let $\Sigma_1=\textbf{B}\textbf{B}^T+\Sigma$ and $\Sigma_2=\textbf{B}\textbf{B}^T$, we have
\begin{eqnarray*}
\log l(x_t,x_s) &=& -\frac{1}{2}\left[\begin{matrix}
    x_t^T & x_s^T
    \end{matrix}\right]\left[\begin{matrix}
    \Sigma_1 & \Sigma_2 \\ \Sigma_2 & \Sigma_1
    \end{matrix}\right]^{-1}\left[\begin{matrix}
    x_t\\ x_s
    \end{matrix}\right]+\frac{1}{2}\left[\begin{matrix}
    x_t^T & x_s^T
    \end{matrix}\right]\left[\begin{matrix}
    \Sigma_1 &0\\0 & \Sigma_1
    \end{matrix}\right]^{-1}\left[\begin{matrix}
    x_t\\ x_s
    \end{matrix}\right]+const. \\
    &=& -\frac{1}{2}\left[\begin{matrix}
   x_t^T & x_s^T
    \end{matrix}\right]\left[\begin{matrix}
    (\Sigma_1 - \Sigma_2\Sigma_1^{-1}\Sigma_2)^{-1} & -\Sigma_1^{-1}\Sigma_2(\Sigma_1 - \Sigma_2\Sigma_1^{-1}\Sigma_2)^{-1} \\ -(\Sigma_1 - \Sigma_2\Sigma_1^{-1}\Sigma_2)^{-1}\Sigma_2\Sigma_1^{-1} & (\Sigma_1 - \Sigma_2\Sigma_1^{-1}\Sigma_2)^{-1}
    \end{matrix}\right]\left[\begin{matrix}
    x_t\\ x_s
    \end{matrix}\right]\nonumber\\&+&\frac{1}{2}\left[\begin{matrix}
   x_t^T & x_s^T
    \end{matrix}\right]\left[\begin{matrix}
    \Sigma_1^{-1} &0\\0 & \Sigma_1^{-1}
    \end{matrix}\right]\left[\begin{matrix}
    x_t\\ x_s
    \end{matrix}\right]+const. \\
    &=& \frac{1}{2}\left[\begin{matrix}
    x_t^T & x_s^T
    \end{matrix}\right]\left[\begin{matrix}
    \Sigma_1^{-1}-(\Sigma_1 - \Sigma_2\Sigma_1^{-1}\Sigma_2)^{-1} & \Sigma_1^{-1}\Sigma_2(\Sigma_1 - \Sigma_2\Sigma_1^{-1}\Sigma_2)^{-1} \\ \Sigma_1^{-1}\Sigma_2(\Sigma_1 - \Sigma_2\Sigma_1^{-1}\Sigma_2)^{-1}& \Sigma_1^{-1}-(\Sigma_1 - \Sigma_2\Sigma_1^{-1}\Sigma_2)^{-1}
    \end{matrix}\right]\left[\begin{matrix}
    x_t\\ x_s
    \end{matrix}\right]+const \nonumber \\
     &=& \frac{1}{2}\left[\begin{matrix}
    x_t^T & x_s^T
    \end{matrix}\right]\left[\begin{matrix}
   \textbf{Q} & \textbf{P} \\ \textbf{P} & \textbf{Q}
    \end{matrix}\right]\left[\begin{matrix}
    x_t\\ x_s
    \end{matrix}\right]+const \nonumber \\
     &=& \frac{1}{2}\left[ x_t^T\textbf{Q}x_t+2x_t^T\textbf{P}x_s+x_s^T\textbf{Q}x_s\right]+const
\end{eqnarray*}

The above approach needs to obtain the inverse of the matrix $\Sigma_1$ and $\Sigma_1 - \Sigma_2\Sigma_1^{-1}\Sigma_2$, both have the size of
$d\times d$. When $d\gg 0$ (that situation is very common in modern application), the inverse becomes very hard to solve or even cannot be solved. If we look carefully about the definition of $\Sigma_1$, it is indeed an inverse of a low-rank correction of $\Sigma$, which can always be computed by doing a low-rank correction to the inverse of the original matrix using the \textbf{Woodbury matrix identity}~\cite{Higham1996Accuracy}:
\begin{equation*}
 \left(A+UCV \right)^{-1} = A^{-1} - A^{-1}U \left(C^{-1}+VA^{-1}U \right)^{-1} VA^{-1},
\end{equation*}
where $A$ is n-by-n and invertible, $U$ is n-by-k, $C$ is k-by-k and invertible, and $V$ is k-by-n.

Using the Woodbury matrix identity, we have
\begin{eqnarray*}
\Sigma_1^{-1}&=&(\Sigma+\textbf{B}\textbf{B}^T)^{-1} \\
&=&\Sigma^{-1} - \Sigma^{-1}\textbf{B} \left(\textbf{I}+\textbf{B}^T\Sigma^{-1}\textbf{B} \right)^{-1} \textbf{B}^T\Sigma^{-1}
\end{eqnarray*}
where we successfully transform the a $d\times d$ matrix inverse into a $N \times N$ matrix inverse.
Using the same identity, with a little cumbersome computation, we have
\begin{eqnarray*}
\left(\Sigma_1 - \Sigma_2\Sigma_1^{-1}\Sigma_2\right)^{-1}&=&\left(\Sigma+\textbf{B}\textbf{B}^T - \textbf{B}\textbf{B}^T\left(\Sigma+\textbf{B}\textbf{B}^T\right)^{-1}\textbf{B}\textbf{B}^T\right)^{-1} \\
&=&\left(\Sigma+\textbf{B}\textbf{B}^T - \textbf{B}\left(\textbf{B}^T\left(\Sigma+\textbf{B}\textbf{B}^T\right)^{-1}\textbf{B}\right)\textbf{B}^T\right)^{-1} \\
&=&\left(\Sigma+ \textbf{B}\left(\textbf{I} - \textbf{B}^T\left(\Sigma+\textbf{B}\textbf{B}^T\right)^{-1}\textbf{B}\right)\textbf{B}^T\right)^{-1} \\
&=&\left(\Sigma+ \textbf{B}\textbf{X}\textbf{B}^T\right)^{-1} \\
&=&\Sigma^{-1} - \Sigma^{-1}\textbf{B} \left(\textbf{X}^{-1}+\textbf{B}^T\Sigma^{-1}\textbf{B} \right)^{-1} \textbf{B}^T\Sigma^{-1}
\end{eqnarray*}
where we define $\textbf{X} = \textbf{I} - \textbf{B}^T\left(\Sigma+\textbf{B}\textbf{B}^T\right)^{-1}\textbf{B}$.

Let $\textbf{Y}=\left(\Sigma_1 - \Sigma_2\Sigma_1^{-1}\Sigma_2\right)^{-1}$ then we have
\begin{eqnarray*}
\log l(x_t,x_s)
    &=& \frac{1}{2}\left[\begin{matrix}
    x_t^T & x_s^T
    \end{matrix}\right]\left[\begin{matrix}
    \Sigma_1^{-1}-(\Sigma_1 - \Sigma_2\Sigma_1^{-1}\Sigma_2)^{-1} & \Sigma_1^{-1}\Sigma_2(\Sigma_1 - \Sigma_2\Sigma_1^{-1}\Sigma_2)^{-1} \\ \Sigma_1^{-1}\Sigma_2(\Sigma_1 - \Sigma_2\Sigma_1^{-1}\Sigma_2)^{-1}& \Sigma_1^{-1}-(\Sigma_1 - \Sigma_2\Sigma_1^{-1}\Sigma_2)^{-1}
    \end{matrix}\right]\left[\begin{matrix}
    x_t\\ x_s
    \end{matrix}\right]+const \\
    &=& \frac{1}{2}\left[\begin{matrix}
    x_t^T & x_s^T
    \end{matrix}\right]\left[\begin{matrix}
    \Sigma_1^{-1}-\textbf{Y} & \Sigma_1^{-1}\Sigma_2\textbf{Y} \\ \Sigma_1^{-1}\Sigma_2\textbf{Y}& \Sigma_1^{-1}-\textbf{Y}
    \end{matrix}\right]\left[\begin{matrix}
    x_t\\ x_s
    \end{matrix}\right]+const \\
    &=&  \frac{1}{2}\left[ x_t^T\left(\Sigma_1^{-1}-\textbf{Y} \right)x_t+x_t^T\left(\Sigma_1^{-1}\Sigma_2\textbf{Y}\right)x_s+x_s^T\left(\Sigma_1^{-1}\Sigma_2\textbf{Y}\right)x_t+x_s^T\left(\Sigma_1^{-1}-\textbf{Y} \right)x_s\right] +const \\
     &=&  \frac{1}{2}\left[ x_t^T\left(\Sigma_1^{-1}-\textbf{Y} \right)x_t+2x_t^T\left(\Sigma_1^{-1}\Sigma_2\textbf{Y}\right)x_s+x_s^T\left(\Sigma_1^{-1}-\textbf{Y} \right)x_s\right]+const \\
       &=&  \frac{1}{2}[ x_t^T\left(- \Sigma^{-1}\textbf{B} \left(\textbf{I}+\textbf{B}^T\Sigma^{-1}\textbf{B} \right)^{-1} \textbf{B}^T\Sigma^{-1} + \Sigma^{-1}\textbf{B} \left(\textbf{X}^{-1}+\textbf{B}^T\Sigma^{-1}\textbf{B} \right)^{-1} \textbf{B}^T\Sigma^{-1} \right)x_t\\
      &+&2x_t^T\left(\Sigma^{-1} - \Sigma^{-1}\textbf{B} \left(\textbf{I}+\textbf{B}^T\Sigma^{-1}\textbf{B} \right)^{-1} \textbf{B}^T\Sigma^{-1}\right)\textbf{B}\textbf{B}^T\left(\Sigma^{-1} - \Sigma^{-1}\textbf{B} \left(\textbf{X}^{-1}+\textbf{B}^T\Sigma^{-1}\textbf{B} \right)^{-1} \textbf{B}^T\Sigma^{-1}\right)x_s\\
      &+&x_s^T\left( - \Sigma^{-1}\textbf{B} \left(\textbf{I}+\textbf{B}^T\Sigma^{-1}\textbf{B} \right)^{-1} \textbf{B}^T\Sigma^{-1} + \Sigma^{-1}\textbf{B} \left(\textbf{X}^{-1}+\textbf{B}^T\Sigma^{-1}\textbf{B} \right)^{-1} \textbf{B}^T\Sigma^{-1} \right)x_s]+const \\
      &=&  \frac{1}{2}[ x_t^T\left(- \Sigma^{-1}\textbf{B}\textbf{T} \textbf{B}^T\Sigma^{-1} + \Sigma^{-1}\textbf{B} \textbf{H} \textbf{B}^T\Sigma^{-1} \right)x_t+x_s^T\left( - \Sigma^{-1}\textbf{B}\textbf{T} \textbf{B}^T\Sigma^{-1} + \Sigma^{-1}\textbf{B} \textbf{H} \textbf{B}^T\Sigma^{-1} \right)x_s\\    &+&2x_t^T\left(\Sigma^{-1}\textbf{B} - \Sigma^{-1}\textbf{B}\textbf{T} \textbf{B}^T\Sigma^{-1}\textbf{B}\right)\left(\textbf{B}^T\Sigma^{-1} - \textbf{B}^T\Sigma^{-1}\textbf{B} \textbf{H} \textbf{B}^T\Sigma^{-1}\right)x_s]+const
\end{eqnarray*}
where we defined $\textbf{T}=\left(\textbf{I}+\textbf{B}^T\Sigma^{-1}\textbf{B} \right)^{-1}$ and $\textbf{H}= \left(\textbf{X}^{-1}+\textbf{B}^T\Sigma^{-1}\textbf{B} \right)^{-1} $ which are inverse of $N\times N$ matrix, since $\Sigma_1$, $\Sigma_2$ and further $\textbf{Y}$ are all symmetric matrix.

\section{Multi-view PLDA}

The goal of this section is to present the EM algorithm updates for leaning the multi-view (joint) PLDA model.

\subsection{Training of multi-view (joint) PLDA}

Let $X=\{x_{ijk}\in \mathbb{R}^d: i=1,...,I;j=1,...,J;k=1,...,H_{ij}\}$ and $x_{ij}=\{x_{ijk}:k=1,...,H_{ij}\}$.

\textbf{E} steps: we need to calculate the expectations $\text{E}_{U|X,\theta_t}[u_i]$, $\text{E}_{V|X,\theta_t}[v_j]$, $\text{E}_{U|X,\theta_t}[u_iu_i^T]$, $\text{E}_{V|X,\theta_t}[v_jv_j^T]$, and  $\text{E}_{U,V|X,\theta_t}[u_iv_j^T]$.
Indeed we have
\begin{eqnarray*}
P(u_i,v_j|x_{ij},\theta)&\propto& P(x_{ij}|u_i,v_j,\theta)P(u_i,v_j) = \left[\prod_{k=1}^{H_{ij}}P(x_{ijk}|u_i,v_j,\theta)\right]P(u_i,v_j)\\
&=&\left[\prod_{k=1}^{H_{ij}}\mathcal{N}(x_{ijk}|\mu+\mathbf{S}u_i+\mathbf{T}v_i, \Sigma)\right]\mathcal{N}(u_i,v_j|0,\textbf{I})\\
&\propto& \left\{\prod_{k=1}^{H_{ij}} \exp \left[-\frac{1}{2}(x_{ijk}-\mu-\mathbf{S}u_i-\mathbf{T}v_j)^T {\Sigma}^{-1}(x_{ijk}-\mu-\mathbf{S}u_i-\mathbf{T}v_j)\right]\right\}\exp\left(-\frac{1}{2}u_i^Tu_i-\frac{1}{2}v_j^Tv_j\right) \nonumber \\
&=& \exp \left\{ -\frac{1}{2}\left[\sum_{k=1}^{H_{ij}} (x_{ijk}-\mu-\textbf{B}z_{ij})^T {\Sigma}^{-1}(x_{ijk}-\mu-\textbf{B}z_{ij})\right] -\frac{1}{2}u_i^Tu_i-\frac{1}{2}v_j^Tv_j\right\} \\
&\propto& \exp \left\{ \sum_{k=1}^{H_{ij}} (x_{ijk}-\mu)^T {\Sigma}^{-1}\textbf{B}z_{ij}-\frac{1}{2}z_{ij}^T(\textbf{I}+\textbf{B}^T{\Sigma}^{-1}\textbf{B})z_{ij}\right\},
\end{eqnarray*}
where $z_{ij}=\left[\begin{matrix}
    u_i \\ v_j
    \end{matrix}\right]$ and $\mathbf{B}=\left[\begin{matrix}
    \mathbf{S} & \mathbf{T}
    \end{matrix}\right]$.

Thus we have
\begin{equation*}
\text{E}_{Z|X,\theta_t}\left\{\left[\begin{matrix}
    u_i \\ v_j
    \end{matrix}\right]\right\}=\text{E}_{Z|X,\theta_t}[z_{ij}]=\left(\textbf{I}+H_{ij}\mathbf{B}^T{\Sigma}^{-1}\mathbf{B}\right)^{-1}\mathbf{B}^T{\Sigma}^{-1}\sum_{k=1}^{H_{ij}}(x_{ijk}-\mu).
\end{equation*}
where $Z=\{z_{ij}: i=1,...,I;j=1,...,J\}$.

\begin{equation*}
\text{E}_{Z|X,\theta_t}\left\{\left[\begin{matrix}
    u_iu_i^T & u_iv_j^T \\ v_ju_i^T & v_jv_j^T
    \end{matrix}\right]\right\}=\text{E}_{Z|X,\theta_t}[z_{ij}z_{ij}^T]=\left(\textbf{I}+H_{ij}\mathbf{B}^T{\Sigma}^{-1}\mathbf{B}\right)^{-1}+\text{E}_{Z|X,\theta_t}[z_{ij}]\text{E}_{Z|X,\theta_t}[z_{ij}]^T.
\end{equation*}

\textbf{M} steps: we update the values of the parameters $\theta=\{\mu, \mathbf{S}, \mathbf{T}, \Sigma\}$ and have
\begin{eqnarray*}
\theta_{t+1} &:=& \arg\max_{\theta}Q(\theta|\theta_t) \\
&=&  \arg\max_{\theta} \text{E}_{U,V|X,\theta_t}[\log \mathcal{L}(\theta;X,U,V)] \\
&=&  \arg\max_{\theta} \text{E}_{U,V|X,\theta_t}[\log p(X,U,V|\theta) ] \\
&=&  \arg\max_{\theta} \text{E}_{U,V|X,\theta_t}\left[\sum_{i=1}^I\sum_{j=1}^{J}\sum_{k=1}^{H_{ij}}\log p(x_{ijk},u_{i},v_{j}|\theta) \right] \\
&=&  \arg\max_{\theta} \text{E}_{U,V|X,\theta_t}\left\{\sum_{i=1}^I\sum_{j=1}^{J}\sum_{k=1}^{H_{ij}}\log [p(x_{ijk}|u_{i},v_{j}, \theta)p(u_i,v_j)] \right\} \\
&=& \arg\max_{\theta} \text{E}_{U,V|X,\theta_t}\left\{\sum_{i=1}^I\sum_{j=1}^{J}\sum_{k=1}^{H_{ij}}\log [ \mathcal{N}(x_{ijk}|\mu + \mathbf{S}u_i+ \mathbf{T}v_j, \Sigma) \mathcal{N}(u_i,v_j|0,\textbf{I})] \right\} \\
&=& \arg\max_{\theta} \text{E}_{U,V|X,\theta_t} \\
& &\left\{\sum_{i=1}^I\sum_{j=1}^{J}\sum_{k=1}^{H_{ij}}\log \left[ \frac{1}{\sqrt{(2\pi)^D|\Sigma|}}\exp \left(-\frac{1}{2}(x_{ijk}-\mu-\mathbf{S}u_i-\mathbf{T}v_i)^T\Sigma^{-1}(x_{ijk}-\mu-\mathbf{S}u_i-\mathbf{T}v_j)\right)\right] \right\}\nonumber \\
&=& - \arg\max_{\theta} \sum_{i=1}^I\sum_{j=1}^{J}\sum_{k=1}^{H_{ij}}\text{E}_{U,V|X,\theta_t}\left\{ \frac{1}{2}\log |\Sigma|+ \frac{1}{2}(x_{ijk}-\mu-\mathbf{S}u_i-\mathbf{T}v_i)^T\Sigma^{-1}(x_{ijk}-\mu-\mathbf{S}u_i-\mathbf{T}v_j)) \right\} \nonumber\\
&=&\arg\max_{\theta} \sum_{i=1}^I\sum_{j=1}^{J}\sum_{k=1}^{H_{ijk}}\left[ - \frac{1}{2}\log |\Sigma| - \frac{1}{2}(x_{ijk}-\mu)^T\Sigma^{-1}(x_{ijk}-\mu)\right] \\& &+ \sum_{i=1}^I\sum_{j=1}^{J}\sum_{k=1}^{H_{ijk}} (x_{ijk}-\mu)^T\Sigma^{-1}\left\{\mathbf{S}\text{E}_{U|X,\theta_t}[u_i]+\mathbf{T}\text{E}_{V|X,\theta_t}[v_i]\right\}
\\ & &- \frac{1}{2}\sum_{i=1}^I\sum_{j=1}^{J}\sum_{k=1}^{H_{ijk}}\left\{\text{E}_{U|X,\theta_t}[u_i^T\mathbf{S}^T\Sigma^{-1}\mathbf{S}u_i]+
2\text{E}_{U,V|X,\theta_t}[v_j^T\mathbf{T}^T\Sigma^{-1}\mathbf{S}u_i]+\text{E}_{V|X,\theta_t}[v_j^T\mathbf{T}^T\Sigma^{-1}\mathbf{T}v_j]\right\} \nonumber
\end{eqnarray*}
In this above derivation, we use the fact that $\mathcal{N}(u_i,v_j|0,\textbf{I})$ has nothing to do with $\theta$.

We have
\begin{equation}
\frac{\partial Q}{\partial \mathbf{S}}=\sum_{i=1}^I\sum_{j=1}^{J}\sum_{k=1}^{H_{ijk}} \Sigma^{-1}(x_{ijk}-\mu)\text{E}_{U|X,\theta_t}[u_i]^T - \sum_{i=1}^I\sum_{j=1}^{J}\sum_{k=1}^{H_{ijk}} \Sigma^{-1}\mathbf{S}\text{E}_{U|X,\theta_t}[u_iu_i^T]-
\sum_{i=1}^I\sum_{j=1}^{J}\sum_{k=1}^{H_{ijk}} \Sigma^{-1}\mathbf{T}\text{E}_{U,V|X,\theta_t}[v_ju_i^T]. \nonumber
\end{equation}
\begin{equation}
\frac{\partial Q}{\partial \mathbf{T}}=\sum_{i=1}^I\sum_{j=1}^{J}\sum_{k=1}^{H_{ijk}}\Sigma^{-1}(x_{ijk}-\mu)\text{E}_{V|X,\theta_t}[v_j]^T - \sum_{i=1}^I\sum_{j=1}^{J}\sum_{k=1}^{H_{ijk}} \Sigma^{-1}\mathbf{T}\text{E}_{V|X,\theta_t}[v_jv_j^T]-
\sum_{i=1}^I\sum_{j=1}^{J}\sum_{k=1}^{H_{ijk}} \Sigma^{-1}\mathbf{S}\text{E}_{U,V|X,\theta_t}[u_iv_j^T]. \nonumber
\end{equation}

Setting $\frac{\partial Q}{\partial \mathbf{S}}=0$ and $\frac{\partial Q}{\partial \mathbf{T}}=0$, we have
\begin{equation*}
\mathbf{S}=\left\{\sum_{i=1}^I\sum_{j=1}^{J}\sum_{k=1}^{H_{ijk}} (x_{ijk}-\mu)\text{E}_{U|X,\theta_t}[u_i]^T-\sum_{i=1}^I\sum_{j=1}^{J}\sum_{k=1}^{H_{ijk}} \mathbf{T}\text{E}_{U,V|X,\theta_t}[v_ju_i^T]\right\}\left\{\text{E}_{U|X,\theta_t}[u_iu_i^T]\right\}^{-1}.
\end{equation*}
\begin{equation*}
\mathbf{T}=\left\{\sum_{i=1}^I\sum_{j=1}^{J}\sum_{k=1}^{H_{ijk}} (x_{ijk}-\mu)\text{E}_{V|X,\theta_t}[v_j]^T-\sum_{i=1}^I\sum_{j=1}^{J}\sum_{k=1}^{H_{ijk}}\mathbf{S}\text{E}_{U,V|X,\theta_t}[u_iv_j^T]\right\}\left\{\text{E}_{V|X,\theta_t}[v_jv_j^T]\right\}^{-1}.
\end{equation*}

We have
\begin{eqnarray*}
\frac{\partial Q}{\partial \Sigma^{-1}} &=&  \frac{1}{2}\sum_{i=1}^I\sum_{j=1}^{J}\sum_{k=1}^{H_{ijk}} \left[\Sigma - (x_{ijk}-\mu)(x_{ijk}-\mu)^T\right] \\
& &+ \sum_{i=1}^I\sum_{j=1}^{J}\sum_{k=1}^{H_{ijk}} (x_{ijk}-\mu)\left[\text{E}_{U|X,\theta_t}[u_i]^T\mathbf{S}^T+\text{E}_{V|X,\theta_t}[v_j]^T\mathbf{T}^T\right] \\ & &- \frac{1}{2}\left[\sum_{i=1}^I\sum_{j=1}^{J}\sum_{k=1}^{H_{ijk}}\left(\mathbf{S}\text{E}_{U|X,\theta_t}[u_iu_i^T]\mathbf{S}^T
+2\mathbf{T}\text{E}_{U,V|X,\theta_t}[v_ju_i^T]\mathbf{S}^T+\mathbf{T}\text{E}_{V|X,\theta_t}[v_jv_j^T]\mathbf{T}^T\right)\right] \\
&=&   \frac{1}{2}\sum_{i=1}^I\sum_{j=1}^{J}\sum_{k=1}^{H_{ijk}} \left[\Sigma - (x_{ijk}-\mu)(x_{ijk}-\mu)^T\right] \\
& &+ \sum_{i=1}^I\sum_{j=1}^{J}\sum_{k=1}^{H_{ijk}} (x_{ijk}-\mu)\left[\text{E}_{U|X,\theta_t}[u_i]^T\mathbf{S}^T+\text{E}_{V|X,\theta_t}[v_j]^T\mathbf{T}^T\right] \\ & &- \frac{1}{2}\left[\sum_{i=1}^I\sum_{j=1}^{J}\sum_{k=1}^{H_{ijk}}\left((x_{ijk}-\mu)\text{E}_{U|X,\theta_t}[u_i]^T\mathbf{S}^T
+(x_{ijk}-\mu)\text{E}_{V|X,\theta_t}[v_j]^T\mathbf{T}^T\right)\right] \nonumber
\end{eqnarray*}

Setting $\frac{\partial Q}{\partial\Sigma^{-1}}=0$, we have
\begin{equation*}
\sum_{i=1}^I\sum_{j=1}^{J}\sum_{k=1}^{H_{ijk}}  \Sigma = \sum_{i=1}^I\sum_{j=1}^{J}\sum_{k=1}^{H_{ijk}}  \left\{(x_{ijk}-\mu)(x_{ijk}-\mu)^T-(x_{ijk}-\mu)\left[\text{E}_{U|X,\theta_t}[u_i]^T\mathbf{S}^T+\text{E}_{V|X,\theta_t}[v_i]^T\mathbf{T}^T\right]\right\}.\nonumber
\end{equation*}

Rearranging, result in
\begin{equation*}
 \Sigma =\frac{1}{\sum_{i=1}^I\sum_{j=1}^{J}\sum_{k=1}^{H_{ijk}}1} \sum_{i=1}^I\sum_{j=1}^{J}\sum_{k=1}^{H_{ijk}} \left\{(x_{ijk}-\mu)(x_{ijk}-\mu)^T-(x_{ijk}-\mu)\left[\text{E}_{U|X,\theta_t}[u_i]^T\mathbf{S}^T+\text{E}_{V|X,\theta_t}[v_i]^T\mathbf{T}^T\right]\right\}. \nonumber
\end{equation*}

We have
\begin{equation*}
\frac{\partial Q}{\partial \mu} = \sum_{i=1}^I\sum_{j=1}^{J}\sum_{k=1}^{H_{ijk}}   - (x_{ijk}-\mu)^T\Sigma^{-1} -\sum_{i=1}^I\sum_{j=1}^{J}\sum_{k=1}^{H_{ijk}} \left[\text{E}_{U|X,\theta_t}[u_i]^T\mathbf{S}^T+\text{E}_{V|X,\theta_t}[v_i]^T\mathbf{T}^T\right]\Sigma^{-1}
\end{equation*}
Setting $\frac{\partial Q}{\partial \mu}=0$, we have
\begin{equation*}
\sum_{i=1}^I\sum_{j=1}^{J}\sum_{k=1}^{H_{ijk}} - (x_{ijk}-\mu)^T- \sum_{i=1}^I\sum_{j=1}^{J}\sum_{k=1}^{H_{ijk}} \left[\text{E}_{U|X,\theta_t}[u_i]^T\mathbf{S}^T+\text{E}_{V|X,\theta_t}[v_j]^T\mathbf{T}^T\right]=0
\end{equation*}

Since $\text{E}_{U|X,\theta_t}[u_i]\approx 0$ and $\text{E}_{V|X,\theta_t}[v_j]\approx 0$, we have $\sum_{i=1}^I\sum_{j=1}^{J}\sum_{k=1}^{H_{ijk}}  - (x_{ij}-\mu)=0$, that is
\begin{equation*}
\mu = \frac{\sum_{i=1}^I\sum_{j=1}^{J}\sum_{k=1}^{H_{ijk}}  x_{ijk}}{\sum_{i=1}^I\sum_{j=1}^{J}\sum_{k=1}^{H_{ijk}}  1}.
\end{equation*}

\subsection{Verification of the latent variables $u$ and $v$ separately}

In multi-view (joint) PLDA, as described in the main body of this work, the latent variables $u$ and $v$ are verified jointly. Indeed the latent variables $u$ and $v$ can also be verified separately.

\begin{figure}[!h]
\centering
\includegraphics[width=0.7\textwidth]{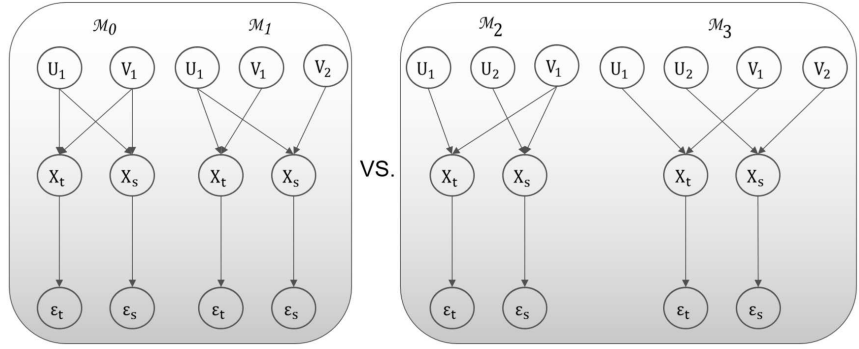}
\caption{Verification by comparing the likelihood of the data under different hypotheses. Under the null hypothesis $\mathcal{H}_0$, the feature $x_t$ and $x_s$ have same latent variable $u$. Under the hypothesis $\mathcal{H}_1$ they have different latent $u$ variables.}
\label{fig:mvPLDAu}       
\end{figure}

Similar with the verification of hidden variables $u$ and $v$ jointly together in the main body, we also treat the verification of single variable $u$ as a kind of hypothesis testing problem with the null hypothesis $\mathcal{H}_0$ defined for the all same latent variable $u$, and the alternative hypothesis $\mathcal{H}_1$ defined for different $u$ variables.
That is in verification we compare the likelihood of the vectors under the hypothesis $\mathcal{H}_0$ where two multi-view features have the same underlying hidden variable $u$ (same $v$ variable in model $\mathcal{M}_0$, different $v$ variables in model $\mathcal{M}_1$, as the Fig.~\ref{fig:mvPLDAu} shows) and the hypothesis $\mathcal{H}_1$ where they have different latent variables of $u$ (same $v$ variable in model $\mathcal{M}_2$, different $v$ variables in model $\mathcal{M}_3$, as the Fig.~\ref{fig:mvPLDA} shows).
Given a test feature $x_t$ and an enrolled feature $x_s$, set $\mathbf{A} = \mathbf{S}\mathbf{S}^T+\mathbf{T}\mathbf{T}^T$, $\mathbf{B} = 2\mathbf{S}\mathbf{S}^T+\mathbf{T}\mathbf{T}^T$, $\mathbf{C} = \mathbf{S}\mathbf{S}^T+2\mathbf{T}\mathbf{T}^T$, and let the priori probability of the models $\mathcal{M}_0$, $\mathcal{M}_1$, $\mathcal{M}_2$, $\mathcal{M}_3$ as $p_0=P(\mathcal{M}_0|\mathcal{H}_0)$, $p_1=P(\mathcal{M}_1|\mathcal{H}_0)$, $p_2=P(\mathcal{M}_2|\mathcal{H}_1)$, $p_3=P(\mathcal{M}_3|\mathcal{H}_1)$, then the likelihood ratio score is

\begin{eqnarray*}
l(x_t,x_s) &=& \frac{P(x_t,x_s|\mathcal{H}_0)}{P(x_t,x_s|\mathcal{H}_1)}\\
&=& \frac{\textbf{X}}{\textbf{Y}}
\end{eqnarray*}
where
\begin{eqnarray*}
& &\textbf{X}\\
&=&P(x_t,x_s|\mathcal{H}_0)\\
&=& P(x_t,x_s|\mathcal{M}_0)P(\mathcal{M}_0|\mathcal{H}_0)+P(x_t,x_s|\mathcal{M}_1)P(\mathcal{M}_1|\mathcal{H}_0)\\
&=&p_0\int\int p(x_t,x_s,u_1,v_1|\theta)du_1adv_1+p_2\int\int\int  p(x_t,x_s,u_1,v_1,v_2|\theta)du_1dv_1dv_2\\
&=&p_0\mathcal{N}(\left[\begin{matrix}
    x_t\\ x_s
    \end{matrix}\right]|\left[\begin{matrix}
    \mu\\ \mu
    \end{matrix}\right],\left[\begin{matrix}
  \mathbf{A}+ \Sigma & \mathbf{A}\\ \mathbf{A}  & \mathbf{A}+\Sigma
    \end{matrix}\right])+p_1\mathcal{N}(\left[\begin{matrix}
    x_t\\ x_s
    \end{matrix}\right]|\left[\begin{matrix}
    \mu\\ \mu
    \end{matrix}\right],\left[\begin{matrix}
    \mathbf{B}  + \Sigma & \mathbf{B}   \\\mathbf{B}   &\mathbf{B}  +\Sigma
    \end{matrix}\right]).
\end{eqnarray*}
and
\begin{eqnarray*}
& &\textbf{Y}\\
&=&P(x_t,x_s|\mathcal{H}_1)\\
&=& P(x_t,x_s|\mathcal{M}_2)P(\mathcal{M}_2|\mathcal{H}_1)+P(x_t,x_s|\mathcal{M}_3)P(\mathcal{M}_3|\mathcal{H}_1)\\
&=&p_2\int\int\int  p(x_t,x_s,u_1,u_2,v_1|\theta)du_1du_2dv_1+p_3\int\int  p(x_t,u_1,v_1|\theta)du_1dv_1\int\int  p(x_s,u_2,v_2|\theta)du_2dv_2\\
&=&p_2\mathcal{N}(\left[\begin{matrix}
    x_t\\ x_s
    \end{matrix}\right]|\left[\begin{matrix}
    \mu\\ \mu
    \end{matrix}\right],\left[\begin{matrix}
  \mathbf{C}+ \Sigma & \mathbf{C}\\ \mathbf{C}  & \mathbf{C}+\Sigma
    \end{matrix}\right])+p_3\mathcal{N}(\left[\begin{matrix}
    x_t\\ x_s
    \end{matrix}\right]|\left[\begin{matrix}
    \mu\\ \mu
    \end{matrix}\right],\left[\begin{matrix}
    \mathbf{A}  + \Sigma & 0  \\ 0  &\mathbf{A}  +\Sigma
    \end{matrix}\right]).
\end{eqnarray*}

\end{document}